\documentclass[11pt, oneside]{article}   	
\usepackage{geometry}                		
\pdfoutput=1						

\usepackage[parfill]{parskip}    			
\usepackage{graphicx}				
\usepackage{amssymb}
\usepackage{amsmath}

\usepackage{bm} 					
\usepackage{setspace}
\usepackage{natbib}					
\setcitestyle{aysep={}}  				
\usepackage[ngerman, english]{babel}
\usepackage{booktabs} 				
\usepackage[table]{xcolor}			
\usepackage{tcolorbox}

\usepackage{multirow} 
\usepackage{arydshln} 
\usepackage{enumitem} 
\setlist{topsep=0pt, leftmargin=*}
\usepackage{float}
\usepackage{titlesec} 
\titlespacing*{\section}
{0pt}{2ex}{0ex}
\titlespacing*{\subsection}
{0pt}{2ex}{0ex}
\titlespacing*{\subsubsection}
{0pt}{2ex}{0ex}

\usepackage{xurl} 
\usepackage[bookmarks]{hyperref}
\hypersetup{colorlinks=true, citecolor=black, linkcolor=black, urlcolor=black} 

\usepackage{sectsty} 
\usepackage[font={sf, small}, labelfont={sf, bf, small}, labelsep=period]{caption}
\allsectionsfont{\sffamily}

\title{{\bfseries\sffamily{Drawing Causal Inferences About \\ Performance Effects in NLP}}}
\author{Sandra Wankm\"uller \\
\emph{Ludwig-Maximilians-Universit\"at M\"unchen} \\
sandra.wankmueller@gsi.lmu.de \\
\small{https://orcid.org/0000-0002-4003-1704}}

\date{}							

\begin{document}
\maketitle

\setlength{\abovedisplayskip}{3pt}
\setlength{\belowdisplayskip}{3pt}

{\bfseries\sffamily{Abstract.}}
This article emphasizes that \emph{NLP as a science} seeks to make inferences about the performance effects that result from applying one method (compared to another method) in the processing of natural language. Yet NLP research in practice usually does not achieve this goal: In NLP research articles, typically only a few models are compared. Each model results from a specific procedural pipeline (here named processing system) that is composed of a specific collection of methods that are used in preprocessing, pretraining, hyperparameter tuning, and training on the target task. To make generalizing inferences about the performance effect that is caused by applying some method $A$ vs.~another method $B$, it is not sufficient to compare a few specific models that are produced by a few specific (probably incomparable) processing systems. Rather, the following procedure would allow drawing inferences about methods' performance effects:
\begin{itemize}
\item A population of processing systems that researchers seek to infer to has to be defined.
\item A random sample of processing systems from this population is drawn. (The drawn processing systems in the sample will vary with regard to the methods they apply along their procedural pipelines and also will vary regarding the compositions of their training and test data sets used for training and evaluation.)
\item Each processing system is applied once with method $A$ and once with method $B$.
\item Based on the sample of applied processing systems, the expected generalization errors of method $A$ and method $B$ are approximated.
\item The difference between the expected generalization errors of method $A$ and method $B$ is the estimated average treatment effect due to applying method $A$ compared to method $B$ in the population of processing systems.
\end{itemize}

\newpage

\section{Drawing Causal Inferences About Performance Effects in NLP}
The actual goal in NLP is to infer how well one method (compared to another method) performs in solving a certain NLP task.\footnote{Often the aim is not only to make inferences with regard to one task but across a range of tasks. But in the following, in order to reduce complexity, the focus will be on one task.} Yet, due to the usual research procedures in NLP, this goal is often not achieved. The common approach in NLP is as follows \citep[p.~1-3]{Reimers2018}: \phantomsection \label{page:pagereference11}
\begin{enumerate}
\item The available annotated data are separated into one training, one validation, and one test set.
\item For each method that is to be compared, a small set of models are trained on the training set and are evaluated on the validation set. 
\item For each method, the model that performs best on the validation set subsequently is evaluated on the test set and the model's performance on the test set is reported.
\end{enumerate}
The described NLP research procedure has two problems.

\subsection{The First Problem}
First, the fact that training is typically conducted on one specific training data set and evaluation is typically conducted on one specific test data set implies that the reported performance values are estimates of the test error 
and not estimates of the expected generalization error. 

The aim of a task in supervised machine learning is to 
approximate the true underlying function $f$ which describes the mapping from inputs $\bm x$ to outputs $y$ for units drawn from joint distribution $p(\bm{x},y)$. When evaluating how well a learning method is able to approximate function $f$, researchers ideally would want to know the expected generalization error that is the expectation of the loss function used for evaluation under the data generating distribution $p(\bm x, y)$:\footnote{Note that for reasons of readability, the notation here does not include parameter values $\bm \theta$. Note furthermore that the term \emph{loss function} $\mathcal L(y_i, \hat{y}_i)$ typically denotes a function that measures the \emph{discrepancy} between true and predicted values. If this is the case, then the loss function really captures an \emph{error} and the smaller the loss value, $L = \mathcal L(y_i, \hat{y}_i)$, the better. However, in the context of evaluation, $\mathcal L(y_i, \hat{y}_i)$ is often a function that measures the \emph{agreement} or \emph{closeness} between true and predicted values. If this is the case, then the higher the value returned by the function, $L = \mathcal L(y_i, \hat{y}_i)$, the better. To consider all loss functions in a consistent framework, in the following, the terms \emph{loss} or \emph{error} are used even if the loss function also can indicate agreement or closeness.}
\begin{equation} \label{eq:ege4}
\mathcal{EGE}(\hat{f}) = \int \int \mathcal L(y, \hat{f}(\bm{x})) p(\bm{x}, y) \, \mathsf{d} \bm{x} \, \mathsf{d} y
\end{equation}
As $p(\bm x, y)$ is unknown, the expected generalization error has to be approximated on the basis of observed data \citep[p.~251-252]{Bischl2012}. In practice, a researcher only has at her disposal a single annotated data set of finite size. She can use one part of the observed data to train a model and then can use another part of the data (then named the test set) to estimate the generalization error of the trained model. The generalization error estimate produced from one such train-test split of the annotated data can be used as an estimate of the expected generalization error. This estimate of the expected generalization error, however, always will depend on the particular composition of the training set and the test set employed \citep[p.~178]{James2013}. In order to produce an estimate of the expected generalization error that takes into account the variability of the loss function value that arises from different train-test set compositions, resampling techniques (e.g.~cross-validation, bootstrapping) can be applied \citep[p.~352]{Bischl2012}. Since resampling techniques are rarely used in NLP, statements about NLP model performances are often based on single train-test splits, and thus are likely to be influenced by idiosyncrasies of the train-test split.

Note that this does not imply that statistical hypothesis tests are generally not conducted when comparing the performances of models in NLP: In some research articles and for some benchmark tasks, statistical hypothesis tests are carried out \citep[p.~1-2]{Reimers2018}. Typically, the null hypothesis is that the performance of two models $A$ and $B$ is equal in the population and the alternative hypothesis is that model performances differ \citep[p.~3]{Reimers2018}. One common way in NLP to implement a statistical hypothesis test is by means of making use of the bootstrap \citep[p.~4]{Reimers2018}. Here, for a given test set $\mathsf{D}_{test}$ of size $M$, the difference in the prediction performance score of model $A$ on $\mathsf{D}_{test}$ and the prediction performance score of model $B$ on $\mathsf{D}_{test}$ is recorded \citep[p.~61]{Riezler2005}. This difference here is indicated by $\delta(\mathsf{D}_{test})$. Then, $\mathsf{K}$ bootstrap samples of size $M$ are drawn at random with replacement from $\mathsf{D}_{test}$ \citep[p.~45]{Efron1993}. On each of the $\mathsf{K}$ bootstrap test sets, the prediction performance of model $A$ and model $B$ is determined and their performance difference $\delta(\mathsf{D}^{\mathsf k}_{test})$ is calculated \citep[p.~61]{Riezler2005}. Hence, one obtains a distribution of the differences in prediction performance values. This bootstrap sampling distribution then is shifted such that it is centered at zero and thus can be used to approximate the distribution of performance differences under the null hypothesis (which assumes that the expectation of performance differences in the population is zero) \citep[p.~61-62]{Riezler2005}. The shifted $\delta(\mathsf{D}^{\mathsf k}_{test})$ here is indicated by $\delta(\mathsf{D}^{\mathsf{k}*}_{test})$. The null hypothesis is rejected if the share of bootstrap test sets for which $\delta(\mathsf{D}^{\mathsf{k}*}_{test}) \geq \delta(\mathsf{D}_{test})$ is smaller than a prespecified significance level (e.g. $\alpha = 0.05$) \citep[p.~61]{Riezler2005}.\footnote{For a general introduction to the bootstrap for hypothesis testing see \citealp{Efron1993} chapter 16.}

This and similar hypothesis tests take into account that the test set is finite. Yet these tests do not take into account that for one test set data point $x^*_m$ the same method when trained on another training set will yield a (slightly) different prediction for $x^*_m$. Bootstrapping on the test set thus is not a proper resampling procedure. The bootstrap sampling distribution of the differences in prediction performance values is not a distribution of the expected generalization error of the learning method. To obtain an adequate estimate of the expected generalization error, a resampling technique in which the method is repeatedly trained on different compositions of the training set and is evaluated on different compositions of the test set is required.

\subsection{The Second Problem}
\label{page:refertohere} This first problem of NLP research procedures can be viewed as a specific instance of the more general second problem. The second problem is that in NLP inference-like statements are often made about methods although models (and not methods) are compared in analyses. This second problem is explicated in the following before the connection between the two problems will be elucidated.

In NLP shared task challenges, the aim is to build a processing system that learns the systematic mapping $f$ from inputs $\bm x$ to outputs $y$ on the basis of a training data set and then makes as accurate as possible predictions for instances in a test set. Therefore, in an NLP challenge, the aim is to build the best performing processing system. But \emph{NLP as a science} seeks to develop---or improve upon---the components of such processing systems. Consequently, \emph{NLP as a science} seeks to infer how one method (vs.~another method) affects prediction performance.

A processing system is not just a learning algorithm, it rather is a procedural pipeline that, for example, may start with pretraining, move on to preprocessing of the target task training documents, and then implement hyperparameter tuning before finally conducting the training process on the target task. A whole collection of methods is involved in implementing such a pipeline. A processing system thus here is conceived of as an object under study that is composed of several methods.

A \emph{method} can be defined at several levels of granularity. Methods that NLP researchers seek to make inferences about can be general and varied entities such as, for example, a learning approach \citep[e.g.][]{Collobert2011}, a model architecture \citep[e.g.][]{Vaswani2017}, or a set of pretraining resources \citep[e.g.][]{Liu2019}, but methods also can be more specific processing elements such as, for example, a pretraining objective \citep[e.g.][]{Devlin2019, Yang2019}, or a hyperparameter setting in fine-tuning \citep[e.g.][]{Mosbach2021}.

When implemented, a processing system produces a trained model $\hat f$ that (more or less well) approximates the true underlying function $f$. Based on an independent test set, function $\mathcal L(y, \hat f(\bm x))$ measures the discrepancy or agreement between the true values $y$ and the values $\hat y =  \hat f(\bm x)$ that are predicted by the trained model. A trained model and its prediction performance as measured by $\mathcal L(y, \hat f(\bm x))$ thus can be understood as the materialized output of a processing system.

Given a population of processing systems of interest, the goal now is to infer the expected value of the effect that using method $A$ compared to method $B$ has on the prediction performance of processing systems in the population. Just as other researchers want to make inferences about the effect of treatment $A$ versus control $B$ in a population of individuals, NLP researchers seek to draw inferences about the effect of treatment method $A$ versus control method $B$ in a population of processing systems. While individuals are characterized by their values on variables, processing systems are characterized by the specific methods they consist of.

In the simple case, a method indeed is like a value on a variable: A processing system consists of numerous elements and each of these elements can be conceived of as a variable $v$ whose values are drawn from a set of methods, e.g.~$v \in \{method \, A, method \, B\}$. For example: One element of a processing system is the pretraining objective. The pretraining objective can be conceived of as a variable that can take on values from a set of methods, which here could be for example~$\{$\emph{language modeling}, \emph{masked language modeling}$\}$. 

Even though a whole processing system, that is composed of various methods, is involved in approximating function $f$, the goal in NLP is to make inferences regarding the effect that one of these methods (compared to another method) has on the performance of the system. In a resource-rich world in which resources are plentiful but not unlimited, the drawing of inferences regarding the performance of one method compared to another method on a given task, could proceed as follows:\footnote{In a world with unlimited resources, drawing inferences about the population would no longer be necessary because the entire population could be analyzed.} \phantomsection \label{page:refertohere1} Two versions of one processing system are constructed. The versions are composed of identical methods, except at one point, where the first version applies method $A$ and the second version implements method $B$. These two versions can be viewed as one individual processing system $\mathsf s$ that is once observed in the treatment group (method $A$) and once in the control group (method $B$). Let $E_{\bm x, y}[\mathcal{L}(y,\hat{f}_{\mathsf s}^{A}(\bm x))]$ be the expectation of the performance of processing system $\mathsf s$ when implementing method $A$ (where the expectation is with respect to the population of data points that are drawn from $p(\bm x, y)$). And let $E_{\bm x, y}[\mathcal{L}(y,\hat{f}_{\mathsf s}^{B}(\bm x))]$ be the expectation of the performance of processing system $\mathsf s$ when implementing method $B$. $E_{\bm x, y}[\mathcal{L}(y,\hat{f}_{\mathsf s}^{A}(\bm x))] - E_{\bm x, y}[\mathcal{L}(y,\hat{f}_{\mathsf s}^{B}(\bm x))]$ then is the individual treatment effect \citep[p.~947]{Holland1986}. The individual treatment effect gives the effect on the expected performance that the treatment of applying method $A$ compared to method $B$ has for individual processing system $\mathsf s$ \citep[p.~947]{Holland1986}. (The fundamental problem of causal inference states that it is not possible to assess 
$E_{\bm x, y}[\mathcal{L}(y,\hat{f}_{\mathsf s}^{A}(\bm x))]$ and $E_{\bm x, y}[\mathcal{L}(y,\hat{f}_{\mathsf s}^{B}(\bm x))]$ on the same entity $\mathsf s$ at the same point in time \citep[p.~947]{Holland1986}. There are several strategies that---when paired with specific assumptions---constitute ways via which the fundamental problem of causal inference can be overcome and inferences can be drawn \citep[p.~947]{Holland1986}. With regard to the individual treatment effect, inference is possible under the assumptions of \emph{temporal stability} and \emph{causal transience} \citep[p.~947]{Holland1986}, which here can be assumed to hold: The values of $E_{\bm x, y}[\mathcal{L}(y,\hat{f}_{\mathsf s}^{A}(\bm x))]$ and $E_{\bm x, y}[\mathcal{L}(y,\hat{f}_{\mathsf s}^{B}(\bm x))]$ do not depend on the particular point in time at which the treatment method $A$ or the control method $B$ are applied. $E_{\bm x, y}[\mathcal{L}(y,\hat{f}_{\mathsf s}^{A}(\bm x))] - E_{\bm x, y}[\mathcal{L}(y,\hat{f}_{\mathsf s}^{B}(\bm x))]$ thus will be constant over time (temporal stability). Moreover, if the two processing system versions are implemented independently of each other without information (e.g.~learned parameters) being able to pass from one system to the other, then the expected performance of processing system $\mathsf s$ when applied with treatment method $A$ will not be affected by previously measuring the expectation of the performance of processing system $\mathsf s$ when applied with method $B$ (causal transience). Hence, $E_{\bm x, y}[\mathcal{L}(y,\hat{f}_{\mathsf s}^{A}(\bm x))] - E_{\bm x, y}[\mathcal{L}(y,\hat{f}_{\mathsf s}^{B}(\bm x))]$ here can be regarded as the causal effect on processing system $\mathsf s$'s expected performance due to the application of treatment method $A$ compared to control method $B$.)

Overall, however, researchers are not interested in the individual treatment effect a method has on a particular processing system $\mathsf s$. Instead, researchers are interested in the average treatment effect of the method in the population of processing systems that a researcher seeks to infer to.\footnote{All units in the population potentially have to be exposable to both compared methods \citep[p.~946]{Holland1986}. For example, if method $A$ is discriminative fine-tuning in which the global learning rate during fine-tuning is different for each layer and if in method $B$ the global learning rate is the same for each layer, then the population of processing systems cannot include conventional machine learning methods that do not have a layered architecture like deep neural networks and thus cannot be exposed to methods $A$ and $B$.} This is, NLP researchers do not seek to make statements like: `For this particular processing system (i.e.~with this particular configuration of methods in pretraining, preprocessing, hyperparameter tuning, and training) method $A$ yields a better expected performance on task $\mathcal T$ than method $B$ by $E_{\bm x, y}[\mathcal{L}(y,\hat{f}_{\mathsf s}^{A}(\bm x))] - E_{\bm x, y}[\mathcal{L}(y,\hat{f}_{\mathsf s}^{B}(\bm x))]$'. Rather, NLP researchers seek to make statements like: `On average method $A$ yields a better expected performance on task $\mathcal T$ than method $B$ by $E_{\mathsf s}[E_{\bm x, y}[\mathcal{L}(y,\hat{f}_{\mathsf s}^{A}(\bm x))] - E_{\bm x, y}[\mathcal{L}(y,\hat{f}_{\mathsf s}^{B}(\bm x))]]$'
(where the expectation is not only with respect to a population of data points but also a population of processing systems) \citep[see also][p.~1-2]{Reimers2018}. This is, in NLP the aim is to draw inferences about methods as tools, that can be plugged into a larger population of processing systems that is of interest when approaching some task. The aim is not to draw inferences about methods as parts of a single, concrete processing system.

As the size of the effect that method $A$ vs.~method $B$ has on the performance is unlikely to be the same across different processing systems in the population (i.e.~there is no unit homogeneity), the individual treatment effect $E_{\bm x, y}[\mathcal{L}(y,\hat{f}_{\mathsf s}^{A}(\bm x))] - E_{\bm x, y}[\mathcal{L}(y,\hat{f}_{\mathsf s}^{B}(\bm x))]$ is not an indicator of the average treatment effect in the population of processing systems \citep[p.~948]{Holland1986}.\footnote{\citet[p.~4]{Assenmacher2021}, for example, show that the performance effect of increasing a model's depth or width is different for different model architectures.} Thus, making inferences regarding the performance of one method vs.~another method in a resource-rich world would imply drawing a random sample of processing systems from the population of processing systems and then applying each processing system once with method $A$ and once with method $B$. The difference between the expectation of the expected value of the performance of the processing systems with method $A$ and the expectation of the expected value of the performance of processing systems with method $B$, $E_{\mathsf s}[E_{\bm x, y}[\mathcal{L}(y,\hat{f}_{\mathsf s}^{A}(\bm x))]] - E_{\mathsf s}[E_{\bm x, y}[\mathcal{L}(y,\hat{f}_{\mathsf s}^{B}(\bm x))]] = E_{\mathsf s}[E_{\bm x, y}[\mathcal{L}(y,\hat{f}_{\mathsf s}^{A}(\bm x))] - E_{\bm x, y}[\mathcal{L}(y,\hat{f}_{\mathsf s}^{B}(\bm x))]]$, then gives an estimate of the average treatment effect in the population \citep[p.~947]{Holland1986}. Subsequently, one could then conduct a hypothesis test with the null hypothesis being that $E_{\mathsf s}[E_{\bm x, y}[\mathcal{L}(y,\hat{f}_{\mathsf s}^{A}(\bm x))] - E_{\bm x, y}[\mathcal{L}(y,\hat{f}_{\mathsf s}^{B}(\bm x))]] = 0$ in the population. Because each processing system is observed once in the treatment and once in the control group, this is a two-dependent-samples problem \citep[p.~210]{Heumann2016}.

An alternative way would be to observe each individual processing system only under method $A$ or under method $B$ instead of in both states. If the processing systems are randomly assigned to the treatment or control group (and thus the assignment to method $A$ or $B$ is independent of other implemented methods of a processing system), then the difference between the average performance value of the processing systems 
in the treatment group, $E_{\mathsf s}[E_{\bm x, y}[\mathcal{L}(y,\hat{f}_{\mathsf s}^{A}(\bm x))]]$, and the average performance value of the processing systems 
in the control group, $E_{\mathsf r}[E_{\bm x, y}[\mathcal{L}(y,\mathtt{f}_{\mathsf r}^{B}(\bm x))]]$, can be used as an estimate of the average treatment effect in the population \citep[p.~948-949]{Holland1986}. \phantomsection In this case one would have a two-independent samples problem \citep[p.~210]{Heumann2016}. \label{page:refertohere2}

In a resource-rich world, the procedure described so far could be implemented to draw inferences about the performance of method $A$ vs.~method $B$ in the population. In the real world with limited resources, the actually implemented NLP research procedures usually differ from what has been described here because of one or both of the following practices. \phantomsection \label{page:nextpageref} \begin{itemize}
\item Two processing systems are compared that not only differ with regard to whether method $A$ or method $B$ is applied but differ with regard to several methods \citep{Assenmacher2020}. Therefore, comparability is not given and it is unclear to which variation of a method a change in performance can be attributed to \citep{Assenmacher2020}.
\item The comparison of applying method $A$ vs.~method $B$ is conducted by incorporating one vs.~the other method in only one (or a few) specific processing systems. Nevertheless, inferences are drawn about the performance of method $A$ vs.~$B$ as general tools within an entire population of processing systems \citep{Reimers2018}. For example: Assume that a processing system is applied once with method $A$ and once with method $B$. Furthermore assume that the performance of the processing system with method $A$ is found to exhibit a higher performance than the processing system with method $B$ and assume that the null hypothesis stating that the performance difference equals zero in the population is rejected on the basis of a hypothesis test that utilizes bootstrapping on the test set. The problematic aspect in NLP research now is that in such a setting the inference is drawn that method $A$ in general (and not only when embedded in this particular processing system) will yield a higher performance than method $B$ \citep{Reimers2018}.
\end{itemize}

When both of these problematic practices are combined in a research procedure, this is as if one would expose one individual to treatment $A$ and would expose another individual, that differs from the first individual with regard to various relevant characteristics, to control $B$ and then would conclude that the higher performance observed for the first individual compared to the second is caused by the treatment of $A$ vs.~$B$ and that thus, in the population in general, applying $A$ compared to $B$ will yield higher performance values.

\section{The First Problem as an Instance of the Second More General Problem and How to Draw Inferences About Performance Effects of Methods}
Up to this point, two problems of NLP research practices have been identified. The first problem is that training and evaluation is often conducted on a single train-test split. Thus, reported performance measures are estimates of the generalization error instead of the expected generalization error. The second problem is that based on the comparison of a few models (that arise from probably incomparable processing systems) conclusions about the causal effect of methods in a larger population of processing systems are drawn.

Above it was stated that the first problem is an instance of the second more general problem (see page \pageref{page:refertohere}). Why is this the case? The question of how a training and a test set is created from a provided annotated data set and thus which data instances the training and the test set are composed of can be regarded as one method within a processing system. The train-test set composition is a characterizing element of a processing system. The processing systems in the population vary regarding the methods that they apply and they vary regarding the train-test set composition they use for training and evaluation. If one processing system were one individual that is observed across a set of variables and if the train-test set composition were one of these variables, then a researcher would like to know what the causal effect of treatment $A$ vs.~control $B$ is not just for an individual with a particular value on the train-test set composition variable. The researcher would like to know what the average effect of treatment $A$ vs.~control $B$ is in a population of individuals, where individuals in this population vary with respect to their values on the observed variables (including the variable of train-test set composition). The problem is that often exactly this knowledge about the population is not generated, because training and evaluation are often conducted on the basis of a single train-test set composition.

One can also pull this up from the other direction: So far, the expected generalization error 
\begin{equation} \label{eq:ege3a}
\mathcal{EGE}(\hat{f}) = \int \int \mathcal L(y, \hat{f}(\bm{x})) p(\bm{x}, y) \, \mathsf{d} \bm{x} \, \mathsf{d} y
\end{equation}
has been defined to be the expectation of the loss function of a learning method in a population of data points that are drawn from the data generating distribution $p(\bm x,y)$. Applying the terminology that is used in this subsection, this definition can be changed as follows: The expected generalization error is the expectation of the loss function of a processing system in a population of data points that are drawn from the data generating distribution $p(\bm x,y)$. Thus, the expected generalization error generalizes over a population of data points (which is what one wants here), but it still is the expected error of a \emph{specific processing system} that is specific to the specific methods that it is composed of. To emphasize this, one can write $\hat f$ in Equation \ref{eq:ege3a} as a specific processing system that results from applying a set of specific methods.

For this purpose, the following assumptions are introduced here: Assume that all processing systems in a population of interest can be described as $\mathsf{g}(v_1, v_2, v_3, v_4, \mathsf{D}_{train}, \mathsf{d}_{test})$. The processing systems in the population consist of four processing elements $v_1, v_2, v_3, v_4$, where each element is a variable 
that can take on values over a set of methods, e.g.~$v_1 \in \{A, B\}; v_2 \in \{C, G, H\}; v_3 \in \{D, I\}; v_4 \in \{F, J, Q, U\}$. The variables here are assumed to be independent and thus the joint probability of a specific method combination, e.g.~$p(v_1 = A, v_2 = C, v_3 = D, v_4 = F)$, is  $p(v_1 = A) p(v_2 = C) p(v_3 = D) p(v_4 = F)$. Also, each variable is assumed to take on each of its methods with equal probability. Thus, e.g.~$p(v_1 = A) = p(v_1 = B)$ and $p(v_1 = A) + p(v_1 = B) = 1$. The joint distribution $p(v_1, v_2, v_3, v_4)$ hence assigns equal probability to each possible combination of methods. Moreover, the processing systems operate on data. A single processing system $\mathsf s$ in this population of processing systems hence is not only characterized by the specific methods it applies but also by the specific data it uses. A processing system can be described as a specific combination of methods and data, e.g.~$\mathsf{g}(v_1 = A, v_2 = C, v_3 = D, v_4 = F, \mathsf{D}^{\mathsf s}_{train}, \mathsf{D}^{\mathsf s}_{test})$. The input to a processing system $\mathsf s$ is a training set of raw data, $\mathsf{D}^{\mathsf s}_{train} = (d_i, y_i)_{i=1}^{N_{\mathsf s}}$, with a given composition and a given size $N_{\mathsf s}$. The methods of a processing system (e.g.~$\{A, C, D, F\}$) process and use the raw training data to produce a trained model $\hat{f}_{\mathsf s}$. $\hat{f}_{\mathsf{s}}$ then can make predictions for instances in the test set, $\mathsf{D}^{\mathsf s}_{test} = (d^{*}_m, y^{*}_m)_{m=1}^{M_{\mathsf s}}$, that likewise has a given composition and size $M_{\mathsf s}$. The instances in the training and test set are assumed to be i.i.d.~samples from $p(d,y)$, which is the joint distribution over raw data $d$ and output values $y$. The performance of the processing system is evaluated by the loss function $\mathcal L$ that compares the true values for the instances in the test set with the values predicted by the system's trained model.

Note that a processing system thus has a data input and a method input. The assumption here is that these input components are mutually independent. (This implies, for example, that whether $v_1 = A$ or $v_1 = B$ will not affect the composition and/or size of $\mathsf{D}^{\mathsf s}_{train}$ and $\mathsf{D}^{\mathsf s}_{test}$.) The methods can be conceived of as functions that process the data and the data can be regarded as the inputs that are being processed.\footnote{Once the system's methods have begun to process the data, independence is no longer given: The parameters and the data representations that the system learns arise as a function of data and methods. But the inputs to the system, so the assumption here, are independent.} The input variables to a processing system, $\{v_1, v_2, v_3, v_4, \mathsf{D}_{train}, \mathsf{D}_{test}\}$, can take values independently without influencing each other. As $\mathsf{D}^{\mathsf s}_{train}$ and $\mathsf{D}^{\mathsf s}_{test}$ are i.i.d.~samples from $p(d,y)$, the joint distribution over input components of processing systems in the population can be described as $p(v_1, v_2, v_3, v_4, d, y)$. The expectation of the loss function in the population of processing systems of interest under raw data distribution $p(d,y)$ thus can be described as\footnote{The representation logic in Equation \ref{eq:ege18} here is as close as possible to the representation logic of Equation  \ref{eq:ege3a}, which is known in the literature. $\sum_{v_1}$ here means the sum over all possible values that $v_1$ can take.}
\begin{equation} \label{eq:ege18}
 \sum_{v_4} \sum_{v_3} \sum_{v_2}  \sum_{v_1} \int \int \mathcal L(y, \mathsf{g}(v_1, v_2, v_3, v_4, d, y)) p(v_1, v_2, v_3, v_4, d, y) \, \mathsf{d} d \, \mathsf{d} y 
\end{equation}
Whereas Equation \ref{eq:ege18} is the expectation of the loss over data and methods, Equation \ref{eq:ege3a}, that is used in the machine learning literature, describes the expected generalization error of an individual processing system that generalizes over data $p(d, y)$ but is specific to the specific combination of methods that it applies. Using the framework introduced here, Equation \ref{eq:ege3a} can be written as:\footnote{Equation \ref{eq:ege3a} uses $p(\bm x, y)$ whereas here---in order to emphasize that it is the processing system that transforms raw data into representations---$p(d, y)$ is used. Equation \ref{eq:ege3c} is the expected generalization error of a processing system with the following combination of methods: $v_1 = A, v_2 = C, v_3 = D, v_4 = F$.} \begin{multline} \label{eq:ege3c}
\mathcal{EGE}(\mathsf{g}(v_1 = A, v_2 = C, v_3 = D, v_4 = F)) = \\ \int \int \mathcal L(y, \mathsf{g}(v_1 = A, v_2 = C, v_3 = D, v_4 = F, d, y)) p(d, y) \, \mathsf{d} d \, \mathsf{d} y
\end{multline}

For NLP researchers, that aim at making inferences about the performance effects of single methods (rather than specific processing systems), the conditional expectation in Equation \ref{eq:ege3c} is not particularly useful. Researchers that seek to estimate the expected loss of applying method $A$ not only in a population of data points but also across a population of processing systems wish to have an estimate for
\begin{equation} \label{eq:ege8}
\mathcal{EGE}(\mathsf{g}(A)) = \sum_{v_4} \sum_{v_3} \sum_{v_2} \int \int \mathcal L(y, \mathsf{g}(v_1 = A, v_2, v_3, v_4, d, y)) p(v_2, v_3, v_4, d, y) \, \mathsf{d} d \, \mathsf{d} y 
\end{equation}
which is the expected generalization error of a method $A$ within a population of processing systems that are trained and evaluated on data points from data generating distribution $p(d, y)$. In contrast to the expected generalization error of a specific processing system presented in Equations \ref{eq:ege3a} and \ref{eq:ege3c}, here all method components (except for method $A$) are averaged over. 

The expected generalization error of method $A$ in Equation \ref{eq:ege8} can be approximated via sampling methods \citep[p.~524]{Bishop2006}: A sample of $\mathsf S$ processing systems is drawn independently from $p(v_2, v_3, v_4, d, y)$. The sampled processing systems vary with regard to the train-test set compositions they operate on and vary with regard to methods $\{v_2, v_3, v_4\}$ that they apply along their procedural pipeline. All processing systems, however, have in common that they apply method $A$. Each sampled processing system is implemented to produce a trained model that then generates predictions for instances in the test set. Then, for each sampled processing system, the loss function value is computed. If $\mathcal L(y^{*}_m, \mathsf{g}(A, v_{2}^{\mathsf s},v_{3}^{\mathsf s},v_{4}^{\mathsf s}, \mathsf{D}^{\mathsf s}_{train}, d^{*}_m))$ denotes the value of the loss function for the $\mathsf s$th sampled processing system on an individual data point from the test set of processing system $\mathsf s$, i.e.~$(d^*_{m}, y^*_{m}) \in \mathsf{D}^{\mathsf s}_{test}$, then Equation \ref{eq:ege8} can be approximated as
\begin{equation}  \label{eq:ege9}
\widehat{\mathcal{EGE}}(\mathsf{g}(A)) = \frac{1}{\mathsf{S}} \sum_{\mathsf{s}=1}^{\mathsf{S}} \frac{1}{|\mathsf{D}^{\mathsf{s}}_{test}|} \sum_{(d^*_{m}, y^*_{m}) \in \mathsf{D}^{\mathsf s}_{test}} \mathcal L(y^{*}_m, \mathsf{g}(A, v_{2}^{\mathsf s},v_{3}^{\mathsf s},v_{4}^{\mathsf s}, \mathsf{D}^{\mathsf s}_{train}, d^{*}_m))
\end{equation}

As described on pages \pageref{page:refertohere1} to \pageref{page:refertohere2} above,\footnote{What has been denoted as $E_{\mathsf s}[E_{\bm x, y}[\mathcal{L}(y,\hat{f}_{\mathsf s}^{A}(\bm x))]]$ on page \pageref{page:refertohere2} above, here is estimated via Equation \ref{eq:ege9}.} given $\widehat{\mathcal{EGE}}(A)$ one can then estimate the causal effect in the population that is due to using method $A$ rather than method $B$ by also applying all sampled processing systems $(1, \dots, \mathsf s, \dots, \mathsf S)$ with method $B$ and then estimating the expected generalization error of method $B$ as 
\begin{equation}  \label{eq:ege10}
\widehat{\mathcal{EGE}}(\mathsf{g}(B)) = \frac{1}{\mathsf{S}} \sum_{\mathsf{s}=1}^{\mathsf{S}} \frac{1}{|\mathsf{D}^{\mathsf{s}}_{test}|} \sum_{(d^*_{m}, y^*_{m}) \in \mathsf{D}^{\mathsf s}_{test}} \mathcal L(y^{*}_m, \mathsf{g}(B, v_{2}^{\mathsf s},v_{3}^{\mathsf s},v_{4}^{\mathsf s}, \mathsf{D}^{\mathsf s}_{train}, d^{*}_m))
\end{equation}
For an individual sampled processing system $\mathsf s$
\begin{multline}
\frac{1}{|\mathsf{D}^{\mathsf{s}}_{test}|} \sum_{(d^*_{m}, y^*_{m}) \in \mathsf{D}^{\mathsf s}_{test}} \mathcal L(y^{*}_m, \mathsf{g}(A, v_{2}^{\mathsf s},v_{3}^{\mathsf s},v_{4}^{\mathsf s}, \mathsf{D}^{\mathsf s}_{train}, d^{*}_m)) \\ - \frac{1}{|\mathsf{D}^{\mathsf{s}}_{test}|} \sum_{(d^*_{m}, y^*_{m}) \in \mathsf{D}^{\mathsf s}_{test}} \mathcal L(y^{*}_m, \mathsf{g}(B, v_{2}^{\mathsf s},v_{3}^{\mathsf s},v_{4}^{\mathsf s}, \mathsf{D}^{\mathsf s}_{train}, d^{*}_m))
\end{multline}
gives an estimate of the individual treatment effect. The difference
\begin{equation}  \label{eq:ege11}
\widehat{\mathcal{EGE}}(\mathsf{g}(A)) - \widehat{\mathcal{EGE}}(\mathsf{g}(B))
\end{equation}
gives an estimate of the average treatment effect in the population.

Alternatively, an estimate of the average treatment effect can be obtained by drawing an i.i.d.~sample of $\mathsf S$ processing systems from $p(v_2, v_3, v_4, d, y)$ and then randomly assigning each processing system in the sample to either treatment method $A$ or control method $B$. Subsequently, $\widehat{\mathcal{EGE}}(\mathsf{g}(A))$ is computed as in Equation \ref{eq:ege9} on the basis of those processing systems that have been assigned to treatment method $A$ and $\widehat{\mathcal{EGE}}(\mathsf{g}(B))$ is computed as in Equation \ref{eq:ege10} using the processing systems assigned to control method $B$. Equation \ref{eq:ege11} then gives an estimate of the average treatment effect. This time, however, $\widehat{\mathcal{EGE}}(\mathsf{g}(B))$ and $\widehat{\mathcal{EGE}}(\mathsf{g}(A))$ come from two independent (rather than two dependent) samples.

Thus far, a situation has been described in which researchers seek to draw inferences about the performance effects of a single, fixed method. There are, however, situations in which researchers want to make inferences about methods in the sense of general and varied entities (e.g.~learning approaches, model architectures, pretraining procedures). In this case, things get a bit more complicated.
A method in this context is not a concrete value of a variable. A method in this context is a set of variables. A method in this more general broader sense can be described as $\bm{\mathcal A} = \{\mathcal{A}_{v_1}, \mathcal{A}_{v_2}, \dots\}$, where $\mathcal{A}_{v_1}, \mathcal{A}_{v_2}, \dots$ are variables and the values of each variable are drawn from a set of concrete method components, e.g.~$\mathcal{A}_{v_1} \in \{A, B\}, \mathcal{A}_{v_2} \in \{C, G, H\}$. Another method $\bm{\mathcal B}$ can be denoted by $\bm{\mathcal B} = \{\mathcal{B}_{v_1}, \mathcal{B}_{v_2}, \dots\}$, where e.g.~$\mathcal{B}_{v_1} \in \{D, I\}, \mathcal{B}_{v_2} \in \{F, Q\}$. Hence, the variables in $\bm{\mathcal B}$ as well as the sets from which the variables in $\bm{\mathcal B}$ draw, need not necessarily be the same as for $\bm{\mathcal A}$. For example, if $\bm{\mathcal A}$ were transfer learning with Transformers and $\bm{\mathcal B}$ were conventional machine learning, then one may have something as $\bm{\mathcal A} = \{\mathcal{A}_{v_1} =$ \emph{pretraining objective, $\mathcal{A}_{v_2} =$ pretraining corpus, $\mathcal{A}_{v_3} =$ Transformer architecture,}$\dots \}$ and e.g.~$\mathcal{A}_{v_1} =$ \emph{pretraining objective} $= \{$\emph{language modeling}, \emph{masked language modeling}$\} \dots$ whereas for $\bm{\mathcal B}$ this could be $\bm{\mathcal B} = \{\mathcal{B}_{v_1} =$ \emph{lowercasing, $\mathcal{B}_{v_2} =$ weighting of elements in document-feature matrix, $\mathcal{B}_{v_3} =$ learning algorithm,}$\dots \}$ and e.g.~$\mathcal{B}_{v_1} =$ \emph{lowercasing} $= \{$\emph{yes}, \emph{no}$\} \dots$ Note that there is a set of possible method component combinations for each method. $\bm{\mathcal A}^{\mathsf s} = \{\mathcal{A}_{v_1} = A, \mathcal{A}_{v_2} = G, \dots\}$ here denotes one possible combination of method components under approach $\bm{\mathcal A}$.

If the population of processing systems of interest is $\mathsf{g}(\mathcal{V}, v_3, v_4, \mathsf{D}_{train}, \mathsf{D}_{test})$, where $\mathcal{V}$ can be method $\bm{\mathcal A}$ or $\bm{\mathcal B}$, i.e.~$\mathcal{V} \in \{\bm{\mathcal A}, \bm{\mathcal B}\}$, and if $\bm{\mathcal A}^{\mathsf s}$ is a specific combination of method components under approach $\bm{\mathcal A}$, then the estimation of the expected generalization error of applying method $\bm{\mathcal A}$ in a population of processing systems under $p(d,y)$ can be achieved by drawing a sample of $\mathsf S$ processing systems from $p(\bm{\mathcal A}, v_3, v_4, d, y)$ and then computing
\begin{equation}  \label{eq:ege92}
\widehat{\mathcal{EGE}}(\mathsf{g}(\bm{\mathcal A})) = \frac{1}{\mathsf{S}} \sum_{\mathsf{s}=1}^{\mathsf{S}} \frac{1}{|\mathsf{D}^{\mathsf{s}}_{test}|} \sum_{(d^*_{m}, y^*_{m}) \in \mathsf{D}^{\mathsf s}_{test}} \mathcal L(y^{*}_m, \mathsf{g}(\bm{\mathcal A}^{\mathsf s},v_{3}^{\mathsf s},v_{4}^{\mathsf s}, \mathsf{D}^{\mathsf s}_{train}, d^{*}_m))
\end{equation}
Hence it is sampled from all possible method component combinations that arise under method $\bm{\mathcal A}$ and then it is averaged also over these combinations.\footnote{Note that the assumption here is that each combination of method components $\bm{\mathcal A}^{\mathsf s}$ has equal probability.} The same procedure can be repeated under method $\bm{\mathcal B}$ and then $\widehat{\mathcal{EGE}}(\mathsf{g}(\bm{\mathcal A})) - \widehat{\mathcal{EGE}}(\mathsf{g}(\bm{\mathcal B}))$ gives an estimate of the average treatment effect of implementing method $\bm{\mathcal A}$ vs.~$\bm{\mathcal B}$ in the population.

The outlined research procedures would allow drawing inferences about the performance effect of method $A$ (or $\bm{\mathcal A}$) compared to method $B$ (or $\bm{\mathcal B}$) in some population of processing systems for some task $\mathcal T$.

One reason why in the field of NLP these outlined procedures are not implemented and instead rather problematic research practices (that have been described on page \pageref{page:nextpageref}) are used, is that resources are limited. The procedures described here require the training of a sample of processing systems, and the sample should have an appropriate size. But already the implementation of one pretraining run usually consumes considerable amounts of resources \citep[see][p.~6]{Assenmacher2020}. And also the training process on the target task can (depending on the model architecture, the data set size, and document lengths) take a substantive amount of time and computational resources.

So, how could research practices be improved given limited resources? The available resources could be allocated better. For example, resources that are used for hyperparameter tuning could be employed more efficiently and be used for resampling instead. From a scientific point of view, knowing the mean performance of a method across ranges of hyperparameter values is arguably more of interest than knowing the performance of a method with the \emph{best performing} hyperparameter setting. This is especially the case with regard to applications in which hyperparameter tuning is followed by training and evaluation on a single train-test split of the labeled data. Instead of training a processing system with tuned hyperparameter values once on a specific training data set and evaluating it once on a specific test data set, it would be more insightful for a processing system (with fixed, non-tuned hyperparameter values) to be trained and evaluated on different train-test set compositions because then an estimate of the expected generalization error of the processing system (see Equation \ref{eq:ege3c}) could be computed.

\section{Conclusion: NLP as a Science}
In general, the field of NLP would benefit if it perceived itself as a science and would move away from engineering the best performing processing system for a task and move toward applying scientific research procedures that allow drawing inferences about the performance effects of methods. Studies that make processing systems comparable and then examine individual treatment effects of single method components \citep[e.g.][]{Assenmacher2021} are an important first step in this direction. A further step then would be to apply the research procedures described here in order to estimate average treatment effects. For this purpose, NLP researchers have to (begin to) see themselves as scientists. They first have to discuss
\begin{itemize}
\item what advances it could bring to the field if they saw themselves as scientists (on the benefits of scientific engineering see e.g.~\citealp[p.~1-5]{Montgomery2012}),
\item what their scientific research interest is (namely drawing causal inferences about the performance effects of methods), and
\item how to operate as scientists (i.e.~what research procedures to use such that causal inferences can be drawn).
\end{itemize}
This discussion process has begun already: In a recent, award-winning article, \citet{Ulmer2022} summarize suggestions for the improvement and adoption of scientific research practices in NLP.

Once this discussion process is over, new research practices are established, and respective studies are planned, a difficulty will be to define populations of processing systems from which the sampling methods draw and that the resulting inferences refer to. Against the background of limited resources, it might be helpful in a concrete application to first define a smaller population than the one actually of interest, and then---as outlined above---to sample from this somewhat smaller population and estimate an average treatment effect accordingly. This would then allow drawing causal inferences at least with respect to a smaller population.

\newpage
\bibliography{Paper03a097_chapterdiss10}
\bibliographystyle{apalike-me2}

\end{document}